\documentclass[11pt]{article}
\usepackage[a4paper]{geometry}
\usepackage[utf8x]{inputenc}
\usepackage{clic2017}
\usepackage{times}
\usepackage{url}
\usepackage{amsmath}
\usepackage{microtype}
\usepackage{url}
\usepackage{amsthm}
\usepackage{adjustbox}
\usepackage{multirow}
\usepackage{multicol}
\usepackage{arydshln}
\usepackage{graphicx}
\usepackage{subcaption}
\usepackage{mathtools}
\usepackage{amssymb}
\usepackage{covington}
\usepackage[T1]{fontenc}

\begin{document}

\author{Severin Laicher, Gioia Baldissin, Enrique Casta\~neda \\ \bf Dominik Schlechtweg, Sabine Schulte im Walde \\
Institute for Natural Language Processing, University of Stuttgart\\
\small{\tt \{laichesn,baldisga,medinaeo,schlecdk,schulte\}@ims.uni-stuttgart.de}\thanks{\ \ ``Copyright \textcopyright\  
2020 for this paper by its authors. Use permitted under Creative Commons License Attribution 4.0 International (CC BY 4.0).''}}  
\date{}

\title{CL-IMS @ DIACR-Ita:\\

Volente o Nolente: BERT does not outperform SGNS on Semantic Change Detection
}

\maketitle

\begin{abstract}
    We present the results of our participation in the DIACR-Ita shared task on lexical semantic change detection for Italian. We exploit Average Pairwise Distance of token-based BERT embeddings between time points and rank 5 (of 8) in the official ranking with an accuracy of $.72$. While we tune parameters on the English data set of SemEval-2020 Task 1 and reach high performance, this does not translate to the Italian DIACR-Ita data set. Our results show that we do not manage to find robust ways to exploit BERT embeddings in lexical semantic change detection.
\end{abstract}

\section{Introduction}
Lexical Semantic Change (LSC) Detection has drawn increasing attention in the past years \cite{kutuzov-etal-2018-diachronic,2018arXiv181106278T}. Recently, SemEval-2020 Task 1 provided a multi-lingual evaluation framework to compare the variety of proposed model architectures \cite{schlechtweg2020semeval}. The DIACR-Ita shared task extends parts of this framework to Italian by providing an Italian data set for SemEval's binary subtask \cite{diacrita_evalita2020}. We present the results of our participation in the DIACR-Ita shared task on lexical semantic change for Italian. We exploit Average Pairwise Distance of token-based BERT embeddings \cite{devlin2018bert} between time points and rank 5 (of 8) in the official ranking with an accuracy of $.72$. While we tune parameters on the English data set of SemEval-2020 Task 1 and reach high performance, this does not transfer to the Italian DIACR-Ita data set. Our results show that we do not manage to find robust ways to exploit BERT embeddings in lexical semantic change detection.

\section{Related Work}

Most existing approaches for LSC detection are type-based \cite{Schlechtwegetal19,Shoemark2019,dubossarskyetal19}. This means that not every word occurrence is considered individually (token-based) but a general vector representation that summarizes every occurrence of a word (including ambiguous words) is created. The results of the SemEval-2020 Task 1 \cite{martinc-etal-2020-context,schlechtweg2020semeval} showed that type-based approaches \cite{prazak-etal-2020-uwb,asgari-etal-2020-emblexchange} achieved better results than token-based approaches \cite{beck-2020-diasense,kutuzov-giulianelli-2020-uiouva}. This is somewhat surprising since in the last years contextualized token-based approaches have achieved significant improvements over the static type-based approaches in several NLP tasks \cite{ethayarajh2019contextual}. \newcite{schlechtweg2020semeval} suggest a range of possible reasons for this: (i) Contextual embeddings are new and lack proper usage conventions. (ii) They are pre-trained and may thus carry additional, and possibly irrelevant, information. (iii) The context of word uses in the SemEval data set was too narrow (one sentence). (iv) The SemEval corpora were lemmatized, while token-based models usually take the raw sentence as input. In the DIACR-Ita challenge (iii) and (iv) are irrelevant because raw corpora with sufficient context are made available to participants. We tried to tackle (i) by excessively tuning parameters and system modules on the English SemEval data set. (ii) can be tackled by fine-tuning BERT on the target corpora. However, our experiments on the English SemEval data set show that  exceptionally high performances can be reached even without fine-tuning.

\section{Experimental setup}
The DIACR-Ita task definition is taken from SemEval-2020 Task 1 Subtask 1 (binary change detection): Given a list of target words and a diacronic corpus pair C$_1$ and C$_2$, the task is to identify the target words which have changed their meanings between the respective time periods t$_1$ and~t$_2$ \cite{diacrita_evalita2020,schlechtweg2020semeval}.\footnote{The time periods t$_1$ and t$_2$ were not disclosed to participants.} C$_1$~and C$_2$ have been extracted from Italian newspapers and books. Target words which have changed their meaning are labeled with the value `1', the remaining target words are labeled with `0'. Gold data for the 18 target words is semi-automatically generated from Italian online dictionaries. According to the gold data, 6 of the 18 target words are subject to semantic change between t$_1$ and t$_2$. This gold data was only made public after the evaluation phase. During the evaluation phase each team was allowed to submit up to 4 predictions for the full list of target words, which were scored using classification accuracy between the predicted labels and the gold data. The final competition ranking compares only the highest of the scores achieved by each team.

\vspace{+3mm}
\section{System Overview}

Our model uses BERT to create token vectors and the average pairwise distance to compare the token vectors from two times. The following chapter presents our model, how we have trained it and how we have chosen our submissions.  

\vspace{+3mm}
\subsection{BERT}
In 2018 Google has released a pre-trained model that ran over Wikipedia and books of different genres \cite{devlin2018bert}: BERT (Bidirectional Encoder Representations from Transformer) is a language representation model, designed to find representations for text by analysing its left and right contexts \cite{devlin2018bert}. \newcite{peters2018dissecting} show that contextual word representations derived from pre-trained bidirectional language models like BERT and ELMo yield significant improvements to the state-of-the-art for a wide range of NLP tasks. BERT can be used to analyse the semantics of individual words, by creating contextualized word representations, vectors that are sensitive to the context in which they appear \cite{ethayarajh2019contextual}. BERT can either create one vector for an input sentence (sentence embedding) or one vector for each input token (token embedding).\footnote{The code of our system is available at \url{https://github.com/Garrafao/TokenChange}.}

Different pre-trained BERT models across languages can be downloaded. In this task, we have used the \textit{bert-base-italian-xxl-cased} model for the Italian language\footnote{\url{https://huggingface.co/dbmdz/bert-base-italian-xxl-cased}} to create token embeddings. 

The basic BERT version is transformer-based and processes text in 12 different layers. In each layer a contextualized token vector representation can be created for each word in an input sentence. It has been claimed that each layer captures different aspects of the input. \newcite{jawahar2019does} suggest that the lower layers capture surface features, the middle layers capture syntactic features and the higher layers capture semantic features of the text. Each layer can serve as representation for the corresponding token by itself, or within a combination of multiple layers.

\vspace{+3mm}
\subsection{Average Pairwise Distance}
Given two sets of token vectors from two time periods t$_1$ and t$_2$, the idea of Average Pairwise Distance (APD) is to randomly pick a number of vectors from both sets and measure their pair-wise distance \cite{Sagi09p104,Schlechtwegetal18,giulianelli2020analysing,beck-2020-diasense,kutuzov-giulianelli-2020-uiouva}. The LSC score of the word is the mean average distance of all comparisons: 

\begin{align*}
\centering
\textnormal{APD}(V,W) = \frac{1}{n_V * n_W} \sum_{v \in V, w \in W} d(v,w)
\end{align*}
where $V$ and $W$ are two sets of vectors, $n_V$ and $n_W$ denote the number of vectors to be compared, and $d(v,w)$ refer to a distance measure (we used cosine distance \cite{SaltonMcGill1983}).  

\vspace{+3mm}
\subsection{Tuning}
The choice of BERT layers and the measure used to compare the resulting vectors (e.g. APD, COS or clustering) strongly influence the performance \cite{kutuzov-giulianelli-2020-uiouva}. Hence, we tuned these parameters/modules on the English SemEval data \cite{schlechtweg2020semeval}. For the 40 English target words we had access to the sentences that were used for the human annotation (in contrast to task participants who had only access to the lemmatized larger corpora containing more target word uses than just the annotated ones).

We tested several change measures regarding their ability to find the actual changing words. As part of our tuning, the APD measure produced the binary and graded LSC scores that best matched the actual LSC scores. We also tested the token vectors from different layers in order to check which one fits best to our task. The best layer combinations were the average of the last four layers and the average of the first and last layer of BERT. The highest F1-score for the binary subtask was $.75$ and a Spearman correlation of $.65$ for the graded subtask. Our results outperformed all official submissions of the shared tasks, of which the best were all type-based. 

\vspace{+3mm}
\subsection{Threshold Selection}
We created four predicted change rankings for the target words with BERT+APD. By experience and consideration of the shared tasks \cite{schlechtweg2020semeval}, we assumed that maximum half of all target words are actual words with a change. Therefore we always annotated at most 9 of 18 words with 1. First, we extracted for each target word a maximum of 200 sentences that contain the word in any token form. We limited the number of uses to 200 for computational efficiency reasons. Then, for each occurrence, we extracted and averaged the token vectors of (i) the last four layers of BERT, and (ii) the first and last layer. For our first submission (`Last Four, 7') we labeled those 7 words with `1' that achieved the highest APD scores in layer combination (i). For our second submission (`First + Last, 7') we labeled those 7 words with `1' that achieved the highest APD scores in layer combination (ii). In (i) and (ii) the same 9 words had the highest APD scores. Therefore, in our third submission (`Average, 9') exactly these 9 words were labeled with `1'. And for our last submission (Lemma, Average, 6') we extracted only sentences in which the target words were present in their lemma form. Again we created the token vectors for the two layer combinations of BERT mentioned above. In both mentioned layer combinations the same 6 words had the highest APD scores. Therefore in our last submission exactly these 6 words were labeled with `1' (similar as in submission 1). 

\vspace{+2mm}
\section{Results}
Table \ref{tab:scoretask2} shows the accuracy scores for the different submissions. 
The best result was achieved by combining the first and last layer of BERT ('First + Last, 7' with $.72$), just like on the SemEval data. The second-best result was obtained by using the sentences where the target word occurred in its lemma form ('Lemma, Average, 6' with $.67$). Only these two submissions outperformed the task baselines and the majority class baseline. The two lowest results were achieved by combining the last four layers of BERT ('Last Four, 7' with $.61$) and by averaging the two layer combinations ('Average, 9' with $.61$). The accuracy of our best submission ($.72$) was ranked at position 5 of the shared task, where the best task result was achieved by two different submissions and reached an accuracy of $.94$. Both submissions were based on type-based embeddings \cite{prazak-etal-2020-CCA,Kaiser2020roots}, clearly outperforming our system.

\vspace{+2mm}
\begin{table}[h]
	\center
%	\small
	\begin{tabular}{ l | r | r}
	\hline
    \textbf{Submission} & \textbf{Thresh.} &   \textbf{Acc.} \\
    \hline
     First + Last &7 & \textbf{.72} \\
	 Lemma, Average & 6 & .67 \\    
 	 \textbf{Majority Class Baseline} &-  & .66 \\
     Average &9 & .61 \\
	 Last Four &7 & .61 \\
	 \textbf{Collocations Baseline} &- & .61 \\
	 \textbf{Frequency Baseline} &-  & .61 \\
    \hline
	\end{tabular}
	\caption{Overview accuracy scores for the four submissions with official task baselines. We also report a majority class baseline of a classifier predicting `0' for all target Words.}
	\label{tab:scoretask2}
\end{table}

\section{Analysis}

As aforementioned, the best performance of our system, achieved with 'First + Last, 7', has an accuracy of $.72$. It erroneously predicts a meaning change for \textit{cappuccio}, \textit{unico} and \textit{campionato}, while for \textit{palmare} and \textit{rampante} it does not detect the change as given by the gold standard. 

We compared both corpora in order to find out if the target words are correctly labeled by the gold standard as well as to identify the possible reasons behind the wrong predictions of our model.

According to our analysis, we can state that the data matches the gold standard.
\textit{Cappuccio} is polysemous across both time periods t$_0$ and t$_1$ (``hood'', ``cap''). However, $31\%$ of the uses in t$_1$ are upper-cased, namely proper nouns (in contrast to the $4\%$ in t$_0$), which might imply a different sense compared to the above-mentioned ones:
\begin{example}
BENEVENTO Il desiderio di il potere , il potere di il desiderio : ruota intorno a questo inquietante ( e attualissimo ) spunto il Festival di Benevento diretto da Ruggero \textbf{Cappuccio} .\\\vspace{2mm}
{\em `BENEVENTO The desire of the power, the power of the desire: the Festival di Benevento directed by Ruggero \textbf{Cappuccio} revolves around this unsettling (and current) cue.'}
\end{example}
This skewed distribution of proper names in the two corpora is a possible reason for the wrong prediction of our model.

Throughout all target words, we noticed that the context provided by the previous and the following sentences (as given as input to our model) is often not related topic-wise; in some instances it seems as if the sentences are headlines, since they refer to different topics:

\begin{example}
M ROMA Sono quindici gli articoli in cui è suddiviso il provvedimento « antiracket » [...]. Roberta Serra ha vinto ieri lo slalom gigante di il \textbf{campionati} italiani femminili . \\\vspace{2mm}
{\em `M ROMA The «antiracket»  measure is divided into fifteen articles [...]. Roberta Serra won yesterday the giant slalom of the Italian female \textbf{championship}.'}
\end{example}
\begin{example}
... le \textbf{uniche} azioni pericolose fiorentine sono arrivate quando il pallone e statu giocato su i lati di il Campo .	costruzione di centrali idroelettriche , di miniere , canali e strade ...\\\vspace{2mm}
{\em `...the \textbf{only} dangerous Florentine actions arrived when the ball  was played on the sides of the field. Construction of hydroelectric power plants, mines, channels and streets...'}
\end{example}

This ``headlines effect'' occurs across the whole corpus. It can be traced back to the extraction process of the original corpus and may be a main source of error in our model. Despite not being representative, the following example shows that in some cases no centric window of any size would avoid considering unrelated context.

\begin{example}
REPARTO CONFEZIONI UOMO GIACCA cameriere bianca , in tessuto L' \textbf{unica} cosa certa è che il governo ha ricevuto una dura lezione da i professori .\\\vspace{2mm}
{\em `MEN'S TAILORING DEPARTMENT white textile waiter JACKET The \textbf{only} certain thing is that the government has received a hard lesson by the professors.'}
\end{example}

\textit{Unico} is  another example of a word that was erroneously predicted as changing. Due to its abstract meaning (``only'', ``single'', ``unique''), it exhibits heterogeneous context across both time periods. Additionally, it can belong to different word classes (noun and adjective in (\ref{ex:noun}) and (\ref{ex:adj}), respectively).

\begin{example}\label{ex:noun}
Rischiamo di rimanere gli \textbf{unici} a non aver dato mano a la ristrutturazione di le Forze Armate .\\\vspace{2mm}
{\em `We risk remaining the \textbf{only ones} not having helped in the reorganization of the Armed Forces.'}
\end{example}
\begin{example}\label{ex:adj}
... è chiaro che l' \textbf{unica} cosa da fare sarebbe l' unificazione di le due aziende comunali ...\\\vspace{2mm}
{\em `...it is clear that the \textbf{only} thing to do would be the unification of the two municipal companies...'}
\end{example}
With regards to the undetected changes, the term \textit{palmare} (polysemous within and across word classes) acquires a novel sense in t$_1$. While it mostly has the meaning of ``evident'' in the 22 sentences of t$_0$ (see (\ref{ex:evident})), it additionally denotes ``palmtop'' in t$_1$ (see (\ref{ex:palmtop})).

\begin{example}\label{ex:evident}
... con evidenza \textbf{palmare} , la impossibilità di difendere una causa perduta ...\\\vspace{2mm}
{\em `with \textbf{undeniable} evidence, the impossibility of defending a lost cause'}
\end{example}%
\begin{example}\label{ex:palmtop}
Per i palestinesi occorre una sistemazione provvisoria in attesa che gli europei si accordino per accoglier li .	Potremmo citare in il lungo elenco il \textbf{palmare} Apple Newton troppo in anticipo su i tempi\\\vspace{2mm}
{\em `A temporary arrangement is needed for the Palestinians while waiting for the Europeans to agree on hosting them. We could quote in the long list the \textbf{palmtop} Apple Newton too far ahead of its time'}
\end{example}
Note that also in (\ref{ex:palmtop}), the topic of the previous and the target sentence is unrelated. 

\textit{Rampante} is a further case of undetected change. The phrase \textit{cavallino rampante}, which metonymically denotes ``Ferrari'', dominates the usage of the word in t$_0$ ($70\%$) and covers a (slightly) relevant share of the uses in t$_1$ ($19\%$). We hypothesize that this leads to a large number of homogenous usage pairs masking the change from ``rampant'', ``unbridled'' to ``extremely ambitious'' of \textit{rampante}.

\section{Conclusion}

Our system comprising BERT+APD was ranked 5 in the DIACR-Ita shared task. The combination of BERT and APD did not perform as well as expected and much lower than the best type-based embeddings, but our best submission still outperformed all baselines. The high tuning results achieved on the SemEval data could not be transferred to the Italian data. One reason for this may be that a different BERT model was applied, trained on text of a different language. We have not tuned the Italian BERT model. It is therefore possible that the decrease in performance may be due to the change of the underlying BERT model. Furthermore, given that our model considers as input also the previous and the following sentences, the presence of semantically unrelated context could have played a significant role in mislabeling the target words.

\section*{Acknowledgments}
Dominik Schlechtweg was supported by the Konrad Adenauer Foundation and the CRETA center funded by the German Ministry for Education and Research (BMBF) during the conduct of this study. We thank the task organizers and reviewers for their efforts.

\bibliographystyle{acl}
\bibliography{biblio}

\end{document}